\documentclass[twoside,11pt]{article} 
\usepackage{jmlr2e}
\usepackage{amsmath,amsfonts,amssymb}
\usepackage{graphicx}
\usepackage{color}
\usepackage{multirow}
 \usepackage{psfrag}
\usepackage{latexsym}
\usepackage{url}
\usepackage{caption}
\usepackage{subcaption}
\usepackage{natbib}
\usepackage{dcolumn}
\usepackage{tikz}
\usepackage{algorithmic}
\usepackage{algorithm}

\usepackage{wrapfig}

\jmlrheading{1}{2000}{1-48}{4/00}{10/00}{Tue Herlau, Morten M{\o}rup, Yee Whye Teh, Mikkel N. Schmidt} 
\ShortHeadings{Adaptive Reconfiguration Moves for Dirichlet Mixtures}{Herlau, M{\o}rup, Teh, Schmidt} %
\firstpageno{1}

\newcommand{\CRP}{\mathrm{CRP}}
\newcommand{\Beta}{\mathrm{Beta}}

\newcommand{\Bernoulli}{\mathrm{Bernoulli}}

\newcommand{\sweep}{ \text{sweep} }

\newcommand{\zstar}[1][ ]{z^{*#1} }

\newcommand{\m}[1]{\boldsymbol{#1}}
\newcommand{\mcal}[1]{\mathcal{#1}}

\newcolumntype{d}[1]{D{.}{.}{#1} }
\usepackage{etoolbox}
\usepackage{booktabs}
\usepackage[detect-all]{siunitx}

\robustify\bfseries%

\makeatletter
\newcommand{\HEADER}[1]{\ALC@it\underline{\textsc{#1}}\begin{ALC@g}}
\newcommand{\ENDHEADER}{\end{ALC@g}}
\makeatother

\newcommand{\STATEI}[1]{\STATE
  \begin{tabular}{@{}p{\dimexpr \textwidth-\labelwidth-\ALC@tlm}@{}}%
    \hangindent \algorithmicindent
    \hangafter 1
    #1
  \end{tabular}
}

\begin{document}

\title{Adaptive Reconfiguration Moves for MCMC inference in Dirichlet Process Mixtures}

\author{\name Tue Herlau \email tuhe@dtu.dk \\
        \name Morten M{\o}rup \email mmor@dtu.dk \\
        \name Mikkel N. Schmidt \email mnsc@dtu.dk \\
       \addr DTU Compute \\
       Technical University of Denmark \\
       Richard Petersens plads 31,
       \\ 2800 Lyngby, Denmark
        \AND
       \name Yee Whye Teh \email y.w.teh@stats.ox.ac.uk \\
       \addr Department of Statistics\\
       University of Oxford \\
       1 South Parks Road, \\
       Oxford OX1 3TG, U.K.}

\editor{Unknown Editor}

\maketitle
\begin{abstract}
Bayesian mixture models are widely applied for unsupervised learning and exploratory data analysis.
Markov chain Monte Carlo based on Gibbs sampling and split-merge moves are widely used for inference in these models. However, both  methods are restricted to limited types of transitions and suffer from torpid mixing and low accept rates even for problems of modest size.
We propose a method that considers a broader range of transitions that are close to equilibrium by
exploiting multiple chains in parallel and using the past states adaptively to inform the proposal distribution.
The method significantly improves on Gibbs and split-merge sampling as quantified using convergence diagnostics and acceptance rates.
Adaptive MCMC methods which use past states to inform the proposal distribution has given rise to many ingenious sampling schemes for continuous problems and the present work can be seen as an important first step in bringing these benefits to partition-based problems. 
\end{abstract}

\section{Introduction}
Mixture models are used for unsupervised learning and exploratory analysis to understand the structure in data by partitioning a set of observations into non-overlapping blocks.
In this work we consider a Bayesian approach to probabilistic inference of partitions where a Dirichlet process is used as the prior for the partitions, and the goal is to estimate the posterior density of partitions using Markov Chain Monte Carlo sampling. Dirichlet process mixture models~\citep{escobar1995bayesian,antoniak1974mixtures,lau2007bayesian} have been applied to a wide range of problems including topic modeling~\citep{teh2004sharing}, multi-task learning for classification~\citep{xue2007multi}, and relational data analysis~\citep{Kemp06,xu2006learning}. 


Two common approaches to inference by MCMC are Gibbs and split-merge (SM) sampling~\citep{neal1991bayesian, jain2004split}. For conjugate models, Gibbs sampling iteratively assigns each observation to a set of candidate blocks corresponding to existing populated blocks or a new block. This makes Gibbs sampling easy to implement. However, as pointed out by~\citet{celeux2000computational} Gibbs sampling is prone to get stuck in local modes and significantly over- or underestimates the true number of blocks. This behavior stem from the incremental nature of Gibbs sampling which prevents the joint movement of observations.

This motivates the use of split-merge operations. Here, 
a split move consist of selecting a single block containing at least two observations and splitting the block into two new blocks thereby increasing the number of blocks by one~\citep{green2001modelling,dahl2003improved,jain2004split}. A merge operation is the inverse procedure where two blocks are merged into one. In the simplest form the proposed split is made at random, however the chance of randomly selecting a favorable split can be very small and these moves will have high reject rate. \cite{jain2004split} proposed using a more complex proposal distribution for a split-move obtained by applying a number of intermediate Gibbs updates to reach equilibrium states. In the following we will use split-merge to refer to the method of \citet*{jain2004split} unless otherwise stated.
\begin{figure}
        \centering
                 \includegraphics[width=\textwidth]{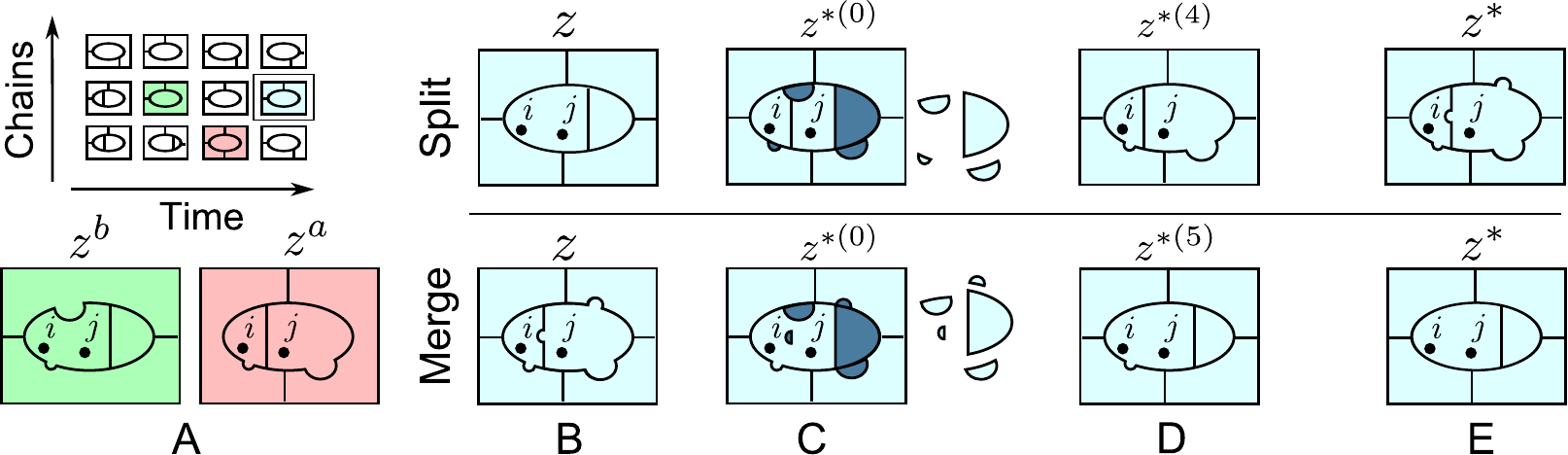}
\caption{A single reconfiguration move applied to the partition $z$. Assume two partitions $z^a$ and $z^b$ as well as two vertices $i,j$ has been chosen from the past states of $S = 3$ chains. In (A) is shown the initial state $z$. by computing the coarsest common refinement between $z,z^a$ and $z^b$ the method construct the initial split and a set of blocks where the 3 partitions agree. These blocks are assumed initially removed from the partition (C). The removed blocks are Gibbs sampled into the problem (D) and finally all singleton elements (subject to certain restrictions) are allowed to move either from outside the blocks containing $i,j$ and into the blocks containing $i,j$, or from inside the blocks containing $i,j$ and out creating the final partition $z^*$ in (E). The bottom row show the same process applied to create the reverse (merge) operation. Notice the total number of blocks in $z,z^*$ remain $6$. }\label{fig:concept}
\end{figure}


While theoretically attractive, split-merge operations have some inherent drawbacks. These include (1) The problem of coupling: split-merge operations restrict themselves to only considering the variables contained in one or two blocks. Often it is the case no single split (or merge) of a block is favorable unless variables from other blocks are allowed to change configuration as well. (2) The problem of poor candidates: Very few blocks of variables should be either split or merged. For this reason, split-merge moves will most of the time attempt to split or merge blocks where the configurations in which they are split (or merged) are highly improbable. (3) The problem of long walks to equilibrium: Split-merge attempt to reach equilibrium states by performing a large update and then slowly sampling towards equilibrium by changing a single variable at a time. While this will eventually reach equilibrium states, the reverse transition probability for a merge move may be very low and it requires many intermediate iterations.

We propose a new method for sampling partition-based models which attempt to address all the above issues (see figure \ref{fig:concept} for an illustration). The primary goal is to overcome the limitation (1) by allowing observations to travel between the blocks currently being split or merged and the other blocks of the partition. For instance figure \ref{fig:concept} illustrates the situation where a subset of a block of variables is being moved from one block to another thereby conserving the total number of blocks. As a result, our method does not require the number of blocks to increase or decrease deterministically, but can in principle change in any direction, this allows for a far larger set of proposal states.

Where split-merge created proposal states near equilibrium through restricted Gibbs sweeps, our method make use of adaptive Markov-Chain Monte Carlo\citep{atchade2005adaptive, roberts2007} to overcome limitation (2) and (3). We evaluate multiple chains in parallel and use disagreement between assignment of vertices between the chains to ensure blocks are only either split or merged if there is disagreement between the chains on their status. In addition, the multiple chains are used to determine the blocks of variables partitions from different chains agree upon and move these blocks jointly, avoiding the many intermediate restricted Gibbs operations in (3).

Due to the adaptive nature of the method and the availability of moves beyond split-merge we dub the method Adaptive Reconfiguration Moves (ARM).

In terms of demands on the model, ARM require the ability to compute change in likelihood when a set of variables are all reassigned at once in a Gibbs sweep. However, for all models we are aware of this change in implementation is a relatively straight-forward generalization of the methods required for Gibbs sampling.

For simplicity we have chosen to focus only on the discrete sampling problems similar to those considered by \cite{jain2004split}, and we will assume the models allow the infinite-dimensional parameters to be integrated out analytically. We evaluate the sampling procedure on two examples of partition-based problems, the Bernoulli mixture model (cf. \citep{everitt1984}) also considered by \cite{jain2004split}, and the Infinite Relational Model of \cite{Kemp06} for undirected graphs.

Sampling operations where blocks are being reassigned also form the basis of the Swendsen-Wang method from statistical physics \cite{swendsen1987nonuniversal}. Swendsen-Wang consider the graph-partitioning problem for two classes and create proposal moves where groups of vertices may change partition by considering disconnected components obtained by stochastically thinning the edges. The generalized Swendsen-Wang algorithm of \cite{barbu2005generalizing} generalize the Swendsen-Wang method to partitions with more than two blocks, however it is still formulated as edge-thinning problem and not obviously amendable to the situation considered herein.

\subsection{Partition-based models}
The Dirichlet process is a prior over measures and is a popular way to induce a prior over random partitions~\citep{ferguson1973bayesian,antoniak1974mixtures}. The Dirichlet process admits several equivalent formulations the most straight-forward of which is the stick-breaking representation of \cite{sethuraman1991constructive}. For a random measure $H$ and concentration parameter $\alpha$ it is given as the generative process
\begin{subequations}\label{eqn:betaconstruct}
\begin{align}
v_k & \sim \Beta(1,\alpha)  & & \theta_k \sim H \\
\beta_k & = v_k \prod_{\ell=1}^{k-1} (1-v_\ell)   & & G = \sum_{k=1}^\infty \beta_k \delta_{\theta_k}
\end{align}
\end{subequations}
which we write $G \sim \textrm{DP}(\alpha,H)$ for the random measure $G$ and $\beta \equiv (\beta_k)_{k=1}^\infty \sim \textrm{GEM}(\alpha)$ for the induced distribution over the weights $\beta_k$~\citep{pitman2002combinatorial}. Consider a set of $n$ observations $\{y_1,\dots,y_n\}$ and assume the following generative model
\begin{subequations}
\label{eqn:DPmix}
\begin{alignat}{2}
 & G & & \sim \textrm{DP}(\alpha,H) \\
\mbox{for $i=1,\dots,n$, }  \quad \theta_i & | G & &  \sim G \\
 y_i & | \theta_i & & \sim F(\theta_i)
\end{alignat}
\end{subequations}
where $F$ is the distribution of $y$ conditional on the parameter $\theta$. The $K$ unique values of $(\theta_i)_{i=1}^n$, denoted $(\theta_k^*)_{k=1}^K$, induce a partition over $[n] = \{1, \dots, n\}$ through the equivalence relation. In particular for each $k = 1, \dots, K$ define $B_k \equiv \{i : \theta_i = \theta^*_k\}$, then the collection of sets $B_k$, written as $z = (B_1, \dots, B_K)$, form a partition of $[n]$ and each $B_k$ is called a \emph{block} of the partition. Since the $\theta_i$ in eq.~\eqref{eqn:betaconstruct} are random, this induces a random partition of $[n]$. The distribution of this random partition is called a \emph{Chinese restaurant process} characterized by $\alpha$ and has the density
\begin{align}
p_{\textrm{CRP}}(z | \alpha) = \frac{\Gamma(\alpha)\alpha^K}{\Gamma(\alpha + n)}\prod_{k=1}^K \Gamma(|B_k|)
\end{align}
where $|B_k|$ denote the number of elements in a block $B_k$, see \citet{pitman2002combinatorial} for addition details.

 We will in the following consider models based on partitions induced by a Dirichlet process and which admit a particular analytical simplification discussed below. In the simulations we will consider vectorial data of the form $y_i$, $i =1,\dots,n$ and relational data consisting of symmetric matrices $y_{ij}$, $1\leq i < j \leq n$, both with representations which may be written as:
\begin{align*}
& & \beta & \sim \textrm{GEM}(\alpha)  &     & &    z^i | \beta & \sim \textrm{Mult}(\beta) & 1 \leq i \leq n \\
\mbox{\emph{Mixture models:} }    & & \theta_k^* & \sim H, & 1\leq k \leq K         &  &  y_i & \sim F(\theta_{z^i}^* ) & 1 \leq i \leq n \\
\mbox{\emph{Relational models:} } & &\theta_{k\ell}^* & \sim H, & 1 \leq k < \ell\leq K  &  &  y_{ij} & \sim F(\theta_{z^i z^j}^* ) & 1\leq i<j \leq n
\end{align*}
where $z^i$ is the index $k$ such that $i \in B_k$ for a partition $z = (B_1,\dots,B_K)$ and with the convention $\theta^*_{\ell k} = \theta^*_{k \ell}$. Notice this representation for the mixture model is equivalent to the generative procedure in eq.~\eqref{eqn:DPmix}. The simplification we will assume is the distributions $F$ and $H$ are conjugated such that all $\theta^*$-parameters can be marginalized out analytically and we will call such a model \emph{conjugate} in the following. Performing this marginalization leave us with a posterior of the form
\begin{align}
p(Y,z) = p(Y | z) p_{\textrm{CRP}}(z) \equiv q(z) \label{eqn:q}
\end{align}
where $Y$ is the data (a vector or matrix). As indicated, we will in the following abbreviate the function $p(Y,z)$ by $q(z)$ and call the above a \emph{model} for partitions.

\subsection{Gibbs sampling}
The methods discussed in this paper are easiest described in notation which admit an ordering of the sets in the partition. Accordingly, the basic object will be a list of sets written $z = (B_1, B_2, \dots, B_K)$ where each $B_\ell$ is denoted a \emph{block} of the partition, however we will in general use the word block to refer to a general subset.

Let $|z|$ define the number of non-empty blocks of $z$ and $\cup z = \cup_{k=1}^{|z|} B_k$ all observations contained in $z$. If each $B_k \neq \emptyset$ and for all $k \neq \ell$: $B_\ell \cap B_k = \emptyset$ we will say $z$ is a \emph{partition} of $X = \cup z$.\\
For a list $z = (B_1,\dots,B_K)$ we let $z(k)$ denote the $k$'th block of $z$, i.e. $z(k) = B_k$. If in addition $z$ is a partition of $X$ and $i \in X$ is any observation, we denote by $z_i$ the unique \emph{block} containing $i$: 
$$
z_i \equiv B_j \text{ such that } i \in B_j
$$
In addition, if $B \in z$ is a block in $z$, denote by $|B|$ the number of elements in $B$. For all $h \leq |B|$ let $B(h)$ indicate the $h$'th value of $B$ in ascending order, specifically $B(1) = \min B$.





We also define common operations on the partitions.
For a block $A$, we let $z \setminus A$ denote the partition $z'$ obtained by removing the set $A$ from each block of $z$ as well as any empty sets. Specifically:
\begin{align*}
z' = \left\{B_k \setminus A: B_k \in z \text{ and } B_k \setminus A \neq \emptyset \right\}
\end{align*}
For two lists of blocks $z^a$, $z^b$ we let $z = z^a \cup z^b$ denote the effect of concatenating $z^b$ after $z^a$. 
Formally let $m = \max\{k : |z^a(k)| > 0\}$ be the last non-empty set of $z^a$, then $z$ is the list such that $z(k) = z^a(k)$ for $k \leq m$ and $z(k) = z^b(k-m)$ for $k > m$.  


Next, for a model of partitions $z$ on a space $X$ (by which we mean a function $q(\cdot)$ as described in eq.~\eqref{eqn:q}) we define the notion of a (restricted) Gibbs sweep of a non-empty block $C \subset X$. A restricted Gibbs sweep is the operation where all variables in the block $C$ are jointly reassigned to a block from a fixed set of available blocks, the particular block being chosen with probability proportional to the likelihood as defined by $q(z)$.

Specifically, for a partition $z = (B_1, B_2, \dots, B_k)$ assume $I$ is a list of blocks $I = (A_1,A_2,\dots,A_m)$ such that for any $A_i$ either $A_i = \emptyset$ or else there exist a $j$ such that $A_i = B_j$. We will also assume $C$, the observations which will be moved, are either contained in a single block of $z$ or disjoint from the observations partitioned by $z$. Formally there either exist $B \in z$ such that $C \subset B$ or $C \cap (\cup z) = \emptyset$. If these conditions are met we define by
\begin{align}
(z^*, \pi_k^*) = \sweep_{C,I}(z) \label{eqn:sweep1}
\end{align}
the operation wherein the observations in $C$ are assigned into one of the blocks in $I$ according to the likelihood. Specifically for each $A_k \in I$ we define a new partition $z^{(k)}$ and weight $\pi_k$ as:
\begin{align*}
z^{(k)} & = (z \setminus (C \cup A_k) ) \cup (C \cup A_k) \\
\pi_k & = \frac{q(z^{(k)}) }{\sum_{m=1}^{|I|} q(z^{(m)})}.
\end{align*}
Notice in non-list notation the partition $z^{(k)}$ is simply
$$
\{B_\ell \setminus (A_k \cup C) : B_\ell \in z \text{ and } B_\ell \setminus (A_k \cup C) \neq \emptyset \} \cup \{A_k \cup C\}
$$
and $z^*$ is obtained as the partition $z^{(k)}$ where 
  $k$ is choosen randomly from the catagorial distribution with weights $(\pi_k)_{k=1}^{|I|}$.
 We will also write
\begin{align}
(z^*, \pi_k)= \sweep_{C,I}(z | z_{C(1)} = k) \label{eqn:sweep2}
\end{align}
as the operation of (forcibly) assigning the variables $C$ into block $k$, i.e., choosing $z^* = z_{(k)}$ and letting $\pi_k$ be the corresponding probability. The standard Gibbs sweeps considered by \cite{maceachern1994estimating} and \cite{neal2000markov} is obtained when $|C| = 1$ and $I$ contain the non-empty blocks of $z$ and an empty set. In order to more easily acommodate this case we use the simplified notation
\begin{align}
(z^*, \pi_k) = \sweep_{C}(z) \label{eqn:sweep3}
\end{align}
for the situation $I = z \cup \{\emptyset\}$ in eqn.~\eqref{eqn:sweep1}.
\subsection{split-merge sampling}
The incremental nature of Gibbs sampling makes it prone to get stuck in local modes where it either over or under estimates the true number of blocks~\citet{celeux2000computational}. Metropolis-Hastings proposal moves are a popular supplement to Gibbs sampling in that it provides a flexible framework to construct bolder update moves based on domain knowledge \cite{metropolis1953equation, hastings1970monte}. The Metropolis-Hastings algorithm samples from a distribution $q$ by first drawing a candidate state $z^*$ according to a proposal density $T_\phi(z^* | z)$ parameterized by $\phi \in \Phi$ and setting the next state of the chain equal to the candidate state with probability
\begin{align}
a(z^*, z) = \min\left[ 1, \frac{T_\phi(z | z^*)}{T_\phi(z^* | z)} \frac{q(z^*)}{q(z)}\right] \label{eqn:acceptrate}
\end{align}
Otherwise the new state remains the current state $z$. The parameters $\phi$ can either be generated deterministically or stochastically and will be discussed later, in the particular case of split-merge it will consist of two observations $i,j$.
\begin{algorithm}
\algsetup{linenosize=\footnotesize}
\footnotesize  
  \begin{algorithmic}[1]
\STATE{
{\bfseries Construct the launch state $z^{(l)}$}: Remove all elements in the block(s) containing $i$ and $j$. If $z_i = z_j$ perform a split otherwise a merge operation. In either case initialize $i$ and $j$ in separate clusters. 
\begin{align}
z^{(l)} &\leftarrow z \setminus \{z_i \cup z_j\},\\
z^{(l)} &\leftarrow z^{(l)} \cup \{ \{i\}, \{j \} \}.
\end{align}
Randomly assign the missing observations $S = z_i \cup z_j \setminus \{i,j\}$ between the two new blocks $z^{(l)}_i,z^{(l)}_j$.}
\STATE{
{\bfseries Perform $L$ restricted Gibbs sweeps on $z^{(l)}$}: Each Gibbs operation iterates over the variables $h \in S$ and sample $h$ between the blocks containing $i$ and $j$
\begin{align}
I & \leftarrow (z^{(l)}_i, z^{(l)}_j), \\
z^{(l)} & \leftarrow \sweep_{h,I}(z^{(l)}).
\end{align}
}
\IF{$z_i \neq z_j$}
    \STATE{
    {\bfseries Merge}: Let $z^*$ be the partition with the two blocks containing $i$ and $j$ merged
    $$
    z^* \leftarrow (z \setminus \{z_i \cup z_j\}) \cup \{z_i \cup z_j\}.
    $$
    Compute the reverse transition probability \emph{from the launch state} by performing one single restricted Gibbs sweep over the observations $h \in S$ forcing the observations to take the same assignment as in $z$
    \begin{align}
    I & \leftarrow (z^{(l)}_i,z^{(l)}_j) \\
    (\cdot, \pi^{h}) & \leftarrow \sweep_{h,I}(z^{(l)} | \mbox{$z^{(l)}_{h} = z^{(l)}_i$ if $z_h = z_i$ otherwise $z^{(l)}_{h} = z^{(l)}_j$} ).
    \end{align}
    Calculate the proposal probability
    $T(z | z^*) = \prod_{h \in S} \pi^h.$
    }
\ELSIF{$z_i = z_j$}
    \STATE{{\bfseries Split}: Otherwise let $z^* = z^{(l)}$ and perform one final restricted sweep over all $h\in S$
    \begin{align}
    I & \leftarrow (\zstar_i,\zstar_j), \\
    (z^*, \pi^{h}) & \leftarrow \sweep_{h,I}(z^* ).
    \end{align}
    Calculate the proposal probability
     $T(z^* | z) = \prod_{h \in S} \pi^h.$
 }
\ENDIF
\end{algorithmic}
\caption{split-merge sampling by \cite{jain2004split}}  \label{alg:SM}
\end{algorithm}

A series of states generated by proposing and accepting according to eq.~\eqref{eqn:acceptrate} will leave the distribution invariant and will sample the problem $q$ provided the chain is ergodic. One particular set of proposal moves is split-merge moves where either a single block is split into two new blocks or two blocks is merged into a single block. While the merge step is unique, there are multiple ways to perform the split step. One of the most popular is the split-merge method of \cite{jain2004split}. The method propose a split configuration by randomly selecting two observations $i,j$ (ie. the background information consist of $\phi = (i,j)$) then randomly distributing the observations assigned to the block(s) containing $i,j$ between two new blocks, then perform a number of Gibbs updates restricted to only moving observations between these two new blocks to obtain near equilibrium split configuration and a final Gibbs update to get a Split-Proposal. It is only the last restricted Gibbs sweep which is used to compute the transition probability $T(z^* | z)$ for a split (in this case the verse probability $T(z | z^*) = 1$) or $T(z | z^*)$ for a merge (in which case $T(z^* | z) = 1$).
Conditional on randomly selected observations $i \neq j$ the construction is listed in algorithm~\ref{alg:SM}.

Convergence of the method can be seen either by considering an augmented target space, or by considering the launch state $z^{(l)}$ as well as the vertices $i,j$ as indices in a very large set of transition kernels which are selected stochastically according to the above procedure \citep{jain2004split,tierney1994markov}. A slight variation of the above method discussed by \cite{dahl2003improved} is obtained by simply omitting the step where the elements are randomly assigned to the blocks of $z^{(l)}$ containing $i$ and $j$, however we have not found significant improvement with this variation for the considered problems.



\section{Proposed Method}
While thermalization of the initial (random) split through restricted Gibbs sweeps can be expected to improve the launch state the method is still limited in two ways: Firstly, based on experimentation with the Infinite Relational Model we found that even when sampling to equilibrium it was often the case there was favorable split or merged configurations of two observations $i,j$, however no single split (or merge) operation of two blocks could reach the more favorable state without altering the assignment of vertices assigned to other blocks.
Secondly, merge moves will tend to have low accept rates unless only a single split configuration is favorable. To illustrate this, suppose the current split configuration is different from that found by performing restricted Gibbs moves on the launch state. Then the final (forced) Gibbs sweep will have to (forcibly) perform a large number of re-assignments to transform the one split-configuration into the other which may often be energetically unfavorable.

\begin{figure}
        \centering
        \begin{subfigure}[b]{0.4\linewidth}
                \raisebox{0.5cm}{\includegraphics[width=\linewidth]{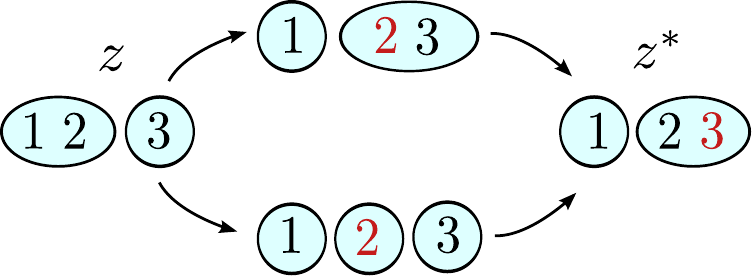}}
                \caption{Uniqueness problem}\label{fig:unique}
        \end{subfigure} %
        ~
        \begin{subfigure}[b]{0.5\linewidth}
                \includegraphics[width=\linewidth]{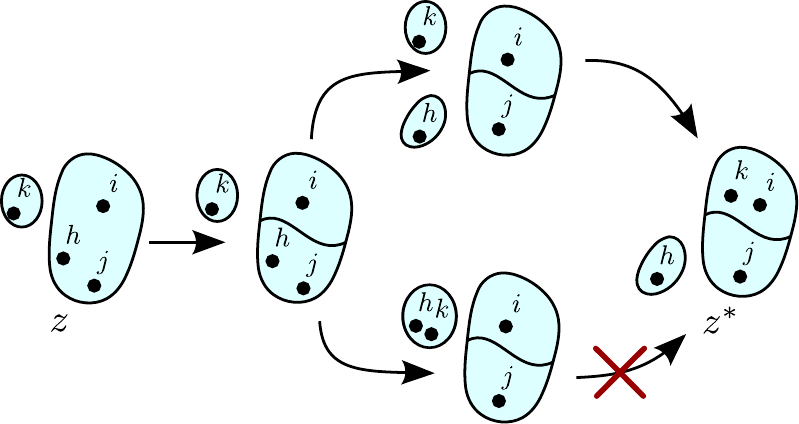}
                \caption{Restricted movement}
        \label{fig:blockpaths}
        \end{subfigure}
\caption{\emph{Left: } Problem defining a unique sampling path for a simple 3-variable problem. The final partition $z^* = \{\{1\}, \{2, 3\}\}$ is obtained from the initial partition $z = \{ \{1,2\}, \{3\}\}$ by Gibbs sampling observation $2,3$. This however allows for two paths and thus the transition probability $T(z|z^*)$ is not the product of each Gibbs transition probability. \emph{Right: } A realistic  uniqueness problem which arise in our method. The observations $h$ and $k$ may arrive at their final position in two ways. To ensure uniqueness, if a block has had variables sampled into it (such as $k$), we disallow all \emph{original} variables to leave the block, thereby removing the bottom path.}
\end{figure}

As outlined in figure \ref{fig:concept}, our proposed method allows observations to not only travel between the two blocks containing $i$ and $j$, but also from blocks not containing $i$ and $j$ and into blocks containing $i$ and $j$ and vice versa. To retain tractability when computing the transition probabilities $T$ we restrict the method to \emph{not} allow observations to travel \emph{between} blocks \emph{not} initially containing $i$ and $j$. The corresponding space of possible transitions is larger than for a Split-Merge operation, and a single restricted Gibbs sweep cannot be expected to produce equilibrium states. We overcome this difficulty by re-using past information of which observations tend to be grouped together and can therefore be expected to change block assignments together. This information is obtained by evaluating multiple chains in parallel and use their agreement or disagreement to both select $i$, $j$ as well as construct the relevant blocks, the idea being that if a set of observations are contained in the same block in both a split and merged configuration it makes sense to update them jointly. Convergence of this scheme is guaranteed under the Adaptive MCMC framework, see \citet[eqs.~(1.1)]{roberts2009examples}.

We will first introduce a simplified version of the method denoted Simplified Reconfiguration Moves (sRM) which does not involve movement of blocks or adaptive MCMC.

\subsection{Simplified Reconfiguration Moves}




Denote by $z$ the current state of the chain and let $z^*$ denote the next state of the chain. Similar to the split-merge algorithm we assume the transition kernels are selected from a set of random kernels $T_\phi(z^* | z)$, $\phi$ again being a set of index parameters.
The method construct $z^*$ through a number of Markov steps $\zstar{(m)}$ indexed by $m$, such that $z^{*(0)}$ is constructed deterministically from the initial state and $\zstar{(M)}$ ($M$ being defined later) corresponds to $z^*$. 

While it is easy to compute the Markov transition probabilities $T_z^{(m)}$ for each step of nearly any construction, the difficulty is to ensure their product corresponds to 
$T(z^* | z)$. The main problem being to ensure the path from $z$ to $z^*$ is unique. Consider for instance the case of a partition of a set of $n=3$ elements, $z = \{ \{1,2\}, \{3\}\}$, and assume a proposal kernel is constructed by first Gibbs sampling element $2$ and then $3$. This does not define a unique path to the final configuration $z^* = \{\{1\}, \{2, 3\}\}$ (see figure~\ref{fig:unique} for an illustration of the two paths) and the transition probability is not simply the product of the transition probability of each Gibbs kernel. For the Split-Merge method uniqueness was ensured since in the final configuration in e.g. a split move, each observation is either in the same block as $i$ or $j$ and the choice is unique, however for the proposed method in which observations can travel between many blocks and blocks can be created or destroyed more care is required.

\begin{algorithm}
 \algsetup{linenosize=\footnotesize}
  \footnotesize
\begin{algorithmic}[1]
  \STATE{
Initialize $T_z \leftarrow 1$.}
\STATE{If $z_i = z_j$ perform a split otherwise merge move.}
\STATE{Remove from $z$ the blocks $z_i$ and $z_j$ containing $i,j$: $\zstar \leftarrow z \setminus \{z_i \cup z_j\}$.}
\STATE{Add $i,j$ to $\zstar$. If splitting:
$\zstar \leftarrow \zstar \cup \{\{i\},\{j\} \}$ else if merging: $\zstar \leftarrow \zstar \cup \{\{i,j\}\}$.
}
\FOR{Each observation $h$ in $\{s : s \neq i,j\}$ } \label{ESMprime:for1}
\IF{$h \in z_i \cup z_j$}
\STATE{Perform an unrestricted Gibbs move of $h$ and update the transition probabilities
$$
 (\zstar,\pi_k)  \leftarrow \sweep_{h}(\zstar) \quad\text{and}\quad T_z \leftarrow \pi_k T_z
$$
}
\ELSIF{$h \notin z_i \cup z_j$}
\IF{$z_h \cap \zstar_h = \{h\}$ and $|\zstar_h| \geq 2$}
\STATE{If other observations has been sampled into $h$'s block in $\zstar$, and $h$ is the last of the original observations from $z$, then do not allow $h$ to change assignment (see eq.~\eqref{eqn:uniquenessconstraint}):
$I \leftarrow (\zstar_h)$.
}
\ELSE
\STATE{Allow $h$ to either stay or move into the block(s) containing $i,j$. \\If splitting: $I \leftarrow (\zstar_i, \zstar_j, \zstar_h)$ otherwise: $I \leftarrow (\zstar_j, \zstar_h)$}
\ENDIF
\STATE{
Perform a Gibbs sweep restricted to the blocks $I$:
$$
 (\zstar,\pi_k)  \leftarrow \sweep_{I,h}(\zstar) \quad\text{and}\quad T_z \leftarrow \pi_k T_z.
$$
}
\ENDIF
\ENDFOR
\end{algorithmic}
\caption{Simplified Reconfiguration Moves}\label{alg:ESMprime}
\end{algorithm}


\subsubsection{Uniqueness of sampling path}
Consider the case outlined in Figure~\ref{fig:blockpaths} corresponding to a split-move of observations $i,j$. In the top-path observation $h$ form a new singleton block and then another singleton block $k$ enter the block containing observation $i$. In the bottom path observation $h$ join the singleton block formed by observation $k$, then $k$ attempt to join the block containing $i$. Since the final partition is the same this creates a non-unique path from $z$ to $z^*$. To disallow this possibility we will impose the restriction if, in the course of a proposal move, a new observation (such as $h$) enters a block, \emph{all} observations originally assigned to this block cannot leave it (such as $k$). Symbolically this restriction corresponds to the case where
\begin{align}
& & \zstar_h \cap z_h & = \{h\}  & & \mbox{ \emph{ ($h$ is the last of the original observations remaining) }} \notag \\
\text{ and } & &  |\zstar_h| & \geq 2 & & \mbox{ \emph{ ($\zstar_h$ contain other observations than the original)} }. \label{eqn:uniquenessconstraint}
\end{align}
This along with the observation that for two blocks $A,B$ \emph{not} containing $i,j$ variables cannot move from $A$ to $B$ or vice versa (see eq.~\eqref{eqn:uniquenessconstraint}) is sufficient to ensure uniqueness. However, we will return to this point after giving the full method. The simplified proposal distribution can be seen as algorithm~\ref{alg:ESMprime}.

\subsection{Full method}
The full method is obtained by including joint updates of blocks of observations in algorithm~\ref{alg:ESMprime} corresponding to plate (C) in figure~\ref{fig:concept}. These blocks are obtained by including, in addition to $i,j$, two partitions $z^a$ and $z^b$ as background information $\phi$ under the restriction $z^a_i = ^a_j$ and $z^b_i \neq z^b_j$. The list of candidate blocks to update jointly are obtained from the coarsest common refinement of $z^a,z^b$ and $z$ \emph{restricted} to the set of observations in the same block as $i,j$, $z_i \cup z_j$. Recall the coarsest common refinement of two partitions $z,z'$ is defined as the partition
$$
z \vee z' = \left\{A \cap B : A \in z, B \in z', A \cap B \neq \emptyset  \right\}.
$$
The intuitive notion is that if variables are assigned similarly in all three blocks then the split (or merge) operation of $i$ and $j$ to obtain the new state $z^*$ will likely leave these consistent assignments invariant as well, see figure~\ref{fig:concept} for an illustration of a single move.

\begin{algorithm}
 \algsetup{linenosize=\footnotesize}
  \footnotesize
  \begin{algorithmic}[1]
  \STATE{
Initialize $T_z \leftarrow 1$ and compute  the coarsest common refinement $c \leftarrow z^a \vee z^b \vee z$.}
\STATE{If $z_i = z_j$ perform a split otherwise a merge move.}
\STATE{Remove from $z$ the blocks $z_i$ and $z_j$ containing $i,j$: $\zstar \leftarrow z \setminus \{z_i \cup z_j\}$.}
\STATE{Add blocks from $c$ containing $i,j$ to $\zstar$. If splitting:
$\zstar \leftarrow \zstar \cup \{c_i, c_j \}$ else if merging: $\zstar \leftarrow \zstar \cup \{c_i \cup c_j \}$.
}
\STATE{Let $g'$ denote all elements of $c$ not currently placed: $g' \leftarrow c \setminus (\cup \zstar)$.}
\FOR{each $a$ in $g'$} \label{ESM:for1}
\STATE{Perform an unrestricted Gibbs move of $a$ and update the transition probabilities
$$
 (\zstar,\pi_k)  \leftarrow \sweep_{a}(\zstar) \quad\text{and}\quad T_z \leftarrow \pi_k T_z.
$$
}
\ENDFOR
\FOR{each observation $h$ not currently updated, ie. in $\{s : s \neq i,j \text{ and } s \neq a(1) \text{ for all } a \in g'\}$ } \label{ESM:for2}
\IF{$h \in z_i \cup z_j$}
\STATE{Perform an unrestricted Gibbs move of $h$ and update the transition probabilities
$$
 (\zstar,\pi_k)  \leftarrow \sweep_{h}(\zstar) \quad\text{and}\quad T_z \leftarrow \pi_k T_z.
$$
}
\ELSIF{$h \notin z_i \cup z_j$}
\IF{$z_h \cap \zstar_h = \{h\}$ and $|\zstar_h| \geq 2$}
\STATE{Implement the uniqueness constraint of eq.~\eqref{eqn:uniquenessconstraint}: $I \leftarrow (\zstar_h)$.}
\ELSE
\STATE{Allow $h$ to either stay or move into the block(s) containing $i,j$. \\If splitting: $I \leftarrow (\zstar_i, \zstar_j, \zstar_h)$ otherwise: $I \leftarrow (\zstar_j, \zstar_h)$}
\ENDIF
\STATE{
Perform a Gibbs sweep restricted to the blocks $I$:
$$
 (\zstar,\pi_k)  \leftarrow \sweep_{I,h}(\zstar) \quad\text{and}\quad T_z \leftarrow \pi_k T_z.
$$
}
\ENDIF
\ENDFOR

\end{algorithmic}
\caption{Adaptive Reconfiguration Move}\label{alg:ESM}
\end{algorithm}

To not restrict the move class, and since the common coarsest refinement of $z,z^a,z^b$ and $z^*,z^a,z^b$ may be different, we allow variables which have been moved as part of a block to be moved independently later. To avoid multiple-path issue we need to ensure each variable can only be updated once. When moving blocks this is ensured by treating the first element of a given block $A$, $A(1)$, as an "earmark" of the block and the other variables $A(2), A(3), \dots$ may then later be updated independently of the rest in a similar fashion as algorithm~\ref{alg:ESMprime}. In other words, when a block $A$ is moved the probabilities in it's Gibbs sweep is computed based on the full likelihood, and since the blocks are constructed to be subsets of $z_i$ and $z_j$ the range of transition probabilities is the full Gibbs move in eq.~\eqref{eqn:sweep1}. However when the other observations of $A$, for instance $A(2)$ is later updated, we compute the available blocks for $A(2)$ \emph{not} based on it's current position (which may be outside $z^*_i$ and $z^*_j$), but again as a full Gibbs sweep since it's original configuration was with $z_i$ and/or $z_j$.


In similar vein to the discussion of \cite{jain2004split, tierney1994markov} we are free to choose the background information $\phi$ deterministically or stochastically. In our approach we will consider a more general setting where $\phi$ depend on the past history of the current chain and other chains, specifically by selecting initial split/merged configurations $z^a$ and $z^b$ which are used to construct the proposal from the past history of the chain. Since these states are selected at random from a growing set of past states, the distribution over pairs will converge and according to the theory of adaptive MCMC we will sample the correct stationary distribution \citep{atchade2005adaptive, roberts2007}. A full description of how the algorithm propose a new configuration $z^*$ and compute the transition probability $T_z$ in eq.~\eqref{eqn:acceptrate} conditional on $i,j,z^a,z^b$ is given in algorithm~\ref{alg:ESM}.




\subsection{Comments on convergence}
We show different paths in the construction of $z^*$ result in different final values of $z^*$, ie. the construction is unique allowing us to identify $T_z$ with the transition probability $T(z^* | z)$. Since the initialization is deterministic the proof proceed by considering each iteration of the for loops in line \ref{ESM:for1} and \ref{ESM:for2} of algorithm~\ref{alg:ESM} and line \ref{ESMprime:for1} of algorithm~\ref{alg:ESMprime} in turn.

For a particular iteration $m$, let $A$ be the block of observations currently being Gibbs sampled. If $A \nsubseteq z_i \cup z_j$ then in the final configuration $z^*$, $A(1)$ (in fact $A$ is a singleton set in this case) will either be the same block as $i$, $j$ or in a different block than both $i$ and $j$. As a result, this branch is unique.

Alternatively, if $A \subseteq z_i \cup z_j$, $A(1)$ may be assigned to a full set of blocks $z^*$ as well as a new block. As before, if $A(1)$ is assigned to the block(s) containing $i$ and $j$ it will remain with $i$ (or $j$) in the final value of $z^*$ making this choice of assignment unique. Accordingly, we only need to consider configurations where $A(1)$ is not assigned to to the block(s) containing $i$ and $j$:

For each such existing candidate block $B_k \in I$, by the non-emptying condition that not all elements of $B_k$ can later be removed: either because they have been assigned to $B_k$ during past iterations of the method (and therefore cannot change assignment later) or if they were in $B_k$ due to their initial assignment in $z$ they cannot all leave due to the non-emptying condition in eq.~\eqref{eqn:uniquenessconstraint}.

In either case there exist elements of each set $B_k$ which are different from $i,j$ and such that $A(1)$ will remain with these elements in the final partition $z^*$ or, if $A$ is assigned to a new block, there is no way for any elements outside of $z_i$ and $z_j$ to end up with $A(1)$. In either case the branch is unique as well.

Finally, to allow a growing number of past states, notice the probability of choosing any two initial states $z^a$, $z^b$ changes proportionally to the inverse of the total number of past states. Since this rate converges to zero, the diminishing adaption condition of \citet[eqs.~(1.1)]{roberts2009examples} is satisfied guaranteeing convergence.



\subsection{Remarks}
To finalize the description of the method we need to specify how the initial information, $z^a,z^b$ and $i,j$ was chosen. Our method evaluated $S$ chains in parallel, such that $Z^{st}$ correspond to the state of chain $s$ at time $t$. When sampling the next state, $Z^{s(t+1)}$, we selected $z^a$ and $z^b$ from the set of $S\lceil t/2 \rceil$ chains
$$
\{Z^{s't'} | s' = 1,\dots,S \text{ and } t' = \lfloor t/2\rfloor, \dots, t\}
$$
at random under the constraints: $z^a \neq z^b$ and one of the chains was selected from the subset where $s' = s$, i.e., the past history of the current chain $s$. Other choices are possible such as using the likelihood of previous states as a weight.

Having selected $z^a$ and $z^b$ we randomly select $i,j$ from the set of all pairs where the two partitions disagreed: $\delta_{z^a_i, z^a_j} \neq \delta_{z^b_i,z^b_j}$ and relabel $z^a$ and $z^b$ if $z^a_i \neq z^a_j$ to agree with our conventions.

The description of ARM and sRM is not fully defined without specifying the order in which the lists of candidate blocks (or variables) are iterated over in line~\ref{ESMprime:for1} of Algorithm~\ref{alg:ESMprime} and line~\ref{ESM:for1} and \ref{ESM:for2} of Algorithm~\ref{alg:ESM}. In the simulations we choose to iterate over the lists according to the size of the blocks (in descending order) and in case of equal size, according to the value of the first element in each block, $A(1)$ (in ascending order). This create a slight dependence on the labelling of the problem and we therefore randomly relabelled the indices of each observation between each iteration.

\begin{wrapfigure}{r}{0.5\textwidth}
\begin{center}
\includegraphics[width=0.9\linewidth]{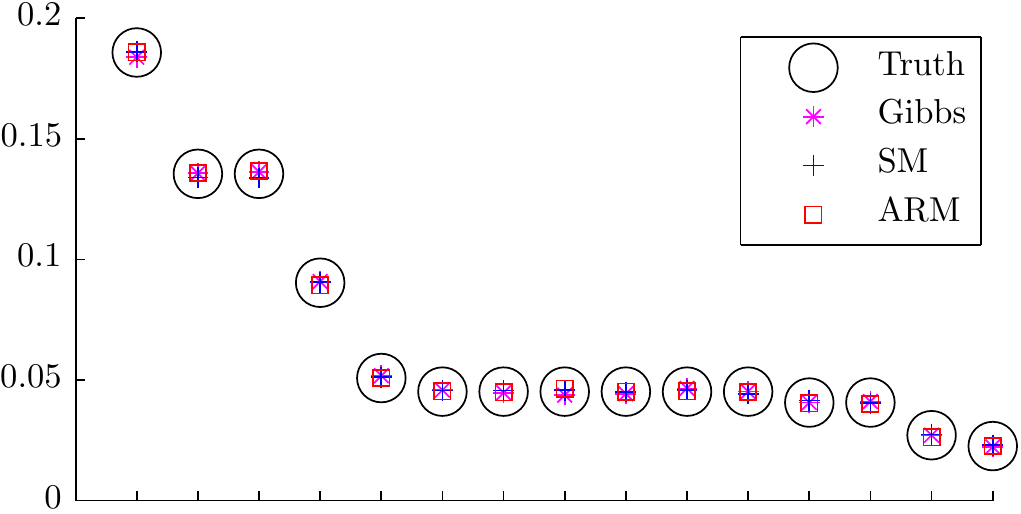}
\end{center}
\caption{Gibbs, SM and ARM (the later two methods not interlaced with Gibbs sweeps) applied to the network problem where $A_{12} = A_{13} = A_{14} = A_{34} = 1$ and otherwise $0$. The plot show the true frequency obtained from evaluating the likelihood and the estimated frequency from the samplers after $160'000$ iterations. All implemented methods recover the true frequency.}\label{fig:sampler_test}
\end{wrapfigure}

\section{Simulations}

Since ARM require initial states $z^a$ and $z^b$ to be well-defined we begin each simulation by evaluating $S$ chains for $50$ iterations using Gibbs sampling to create an initial value of $Z$. To avoid any unfair advantage this initialization was used for all methods. In the simulations, both ARM and SM sampling was interlaced with Gibbs sweeps. Ie. a single iteration consist of a SM or AR move followed by a full Gibbs sweep where each observation $i=1,\dots,n$ is updated according to eq.~\eqref{eqn:sweep3}. The number of intermediate (restricted) Gibbs sweeps for SM sampling was set to $L=5$ in all experiments.

To evaluate the method under diverse and realistic conditions we examine a relational and mixture-type model. The first is the Bernoulli mixture model and the artificial data of the same type considered in \cite{jain2004split}, the second is the Infinite Relational Model applied to four realistic datasets. To evaluate the correctness of the method we first ran each of the 3 methods, Gibbs sampling, SM and ARM on an Infinite Relational Model described in section \ref{sec:IRMsimulations} applied to a simple network problem with 4 observations (vertices) and 6 edges giving a total of $16$ partitions. We compared the empirical frequency obtained from the samplers (without interlacing with Gibbs sweeps for the SM and ARM samplers) over $160'000$ iterations with the true value obtained by evaluating the likelihood and normalizing. The results can be seen in figure~\ref{fig:sampler_test}. Since the samplers are correlated the problem is not easily amendable to standard statistical tests, however the frequency obtained by the three methods and the true frequency obtained from the likelihood are visually in good agreement.

\subsection{Artificial Data}
The Bernoulli Mixture Model for a $d$ feature $\times$ $n$ observations matrix $A$ corresponds to the generative process and likelihood term of the form
\begin{align*}
z & \sim \CRP(\alpha), &   \theta_{ik} & \sim \Beta(\beta_0^+, \beta_0^-), \ i=1,\dots,d, k=1,2,\dots \\
A_{ij} & \sim \Bernoulli(\theta_{iz_j}) &
\log p(A | z) & = \sum_{\substack{k = 1, \dots, K \\ i=1,\dots,d} } \log  \frac{B(N^+_{ik} + \beta_0^+, N^-_{ik} + \beta_0^-)}{B(\beta_0^+,\beta_0^-)}
\end{align*}
where
$N^+_{ik} = \sum_{j \in z(k)} A_{ij}$, $N^-_{ik} = \sum_{j \in z(k)}(1-A_{ij})$ and $\beta_0^+ = \beta_0^- = \alpha = 1$.
We generated artificial data as described by \cite{jain2004split}. The data was composed by dividing $n=100$ observations into $K=5$ components each of size 20. Each component had a variable number $d$ of attributes such that the probability an observation assigned to a component $k$ would have a particular attribute is given in table~\ref{table:njdat}.

\begin{table}
\centering
\begin{tabular}{c l l l l l l}
\toprule
  k & \multicolumn{6}{c}{ $ p(A_{ij} = 1 | z_j = k), \ i=1,\dots,6$ }  \\  \midrule
  1 & .95 & .95 & .95 & .95 & .95 & .95 \\
  2 & .05 & .05 & .05 & .05 & .95 & .95 \\
  3 & .95 & .05 & .05 & .95 & .95 & .95 \\
  4 & .05 & .05 & .05 & .05 & .05 & .05 \\
  5 & .95 & .95 & .95 & .95 & .05 & .05 \\
  \bottomrule
\end{tabular}
\caption{Mixture proportions of the Bernoulli Mixture model. For problems with more than six attributes the last column is simply copied the remaining number of times to form 3 problems with either $d=6, 8$ or $10$ attributes.}\label{table:njdat}
\end{table}

\begin{figure}
        \centering
        \begin{subfigure}[b]{0.3\textwidth}
                \includegraphics[width=\textwidth]{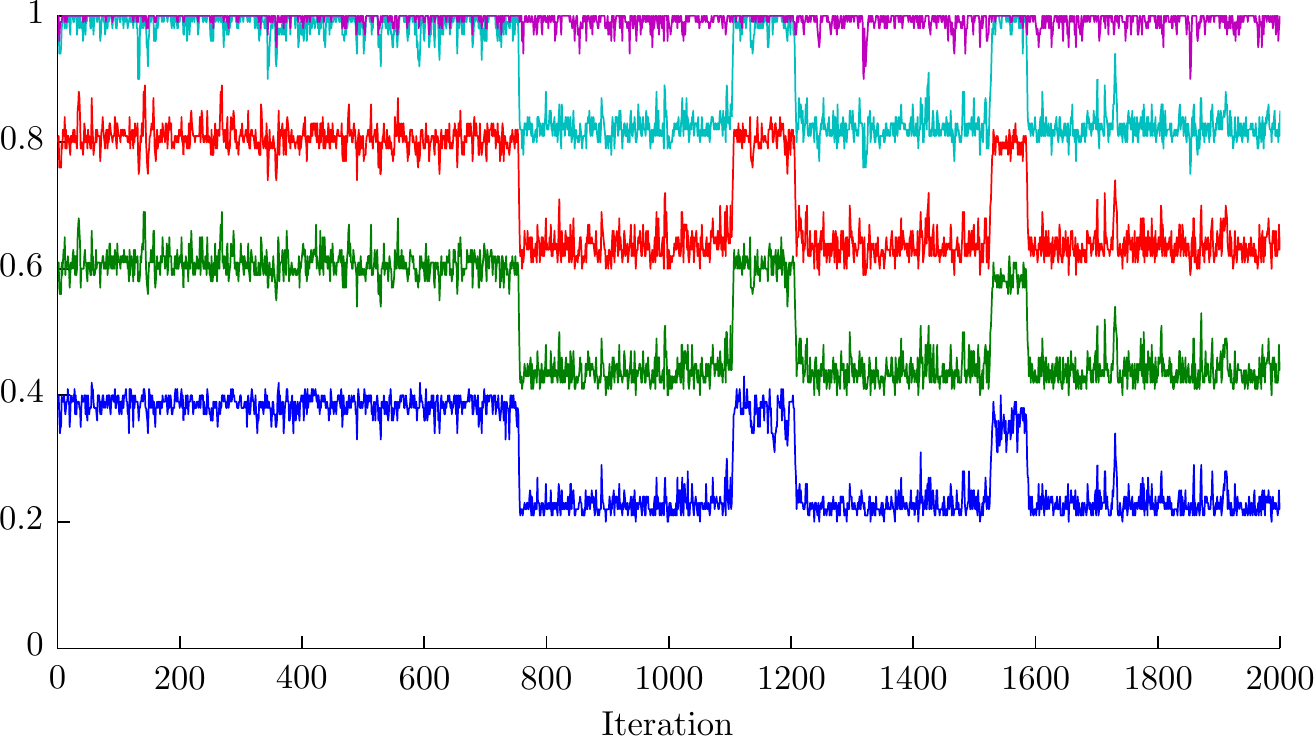}
                \caption{Gibbs sampling}
                \label{fig:btrace1}
        \end{subfigure}%
        ~ 
        \begin{subfigure}[b]{0.3\textwidth}
                \includegraphics[width=\textwidth]{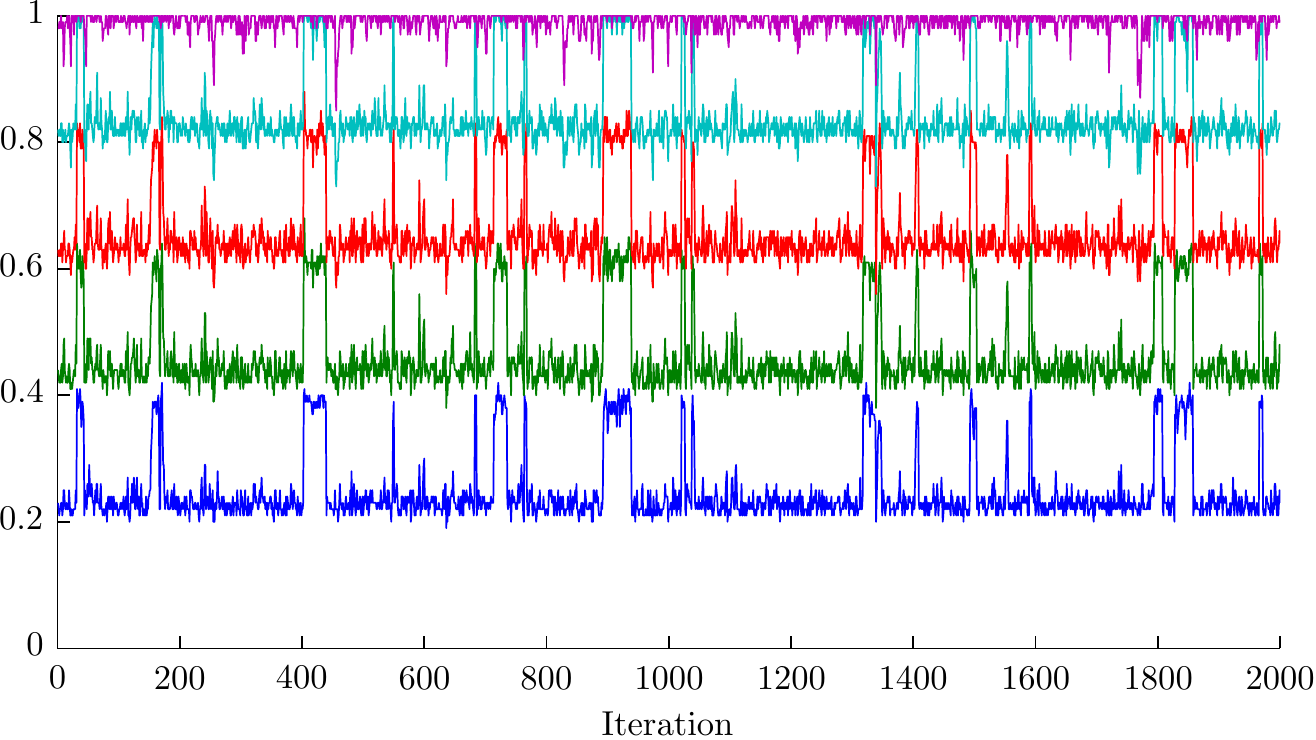}
                \caption{GSM}
                \label{fig:btrace2}
        \end{subfigure}
        ~ 
        \begin{subfigure}[b]{0.3\textwidth}
                \includegraphics[width=\textwidth]{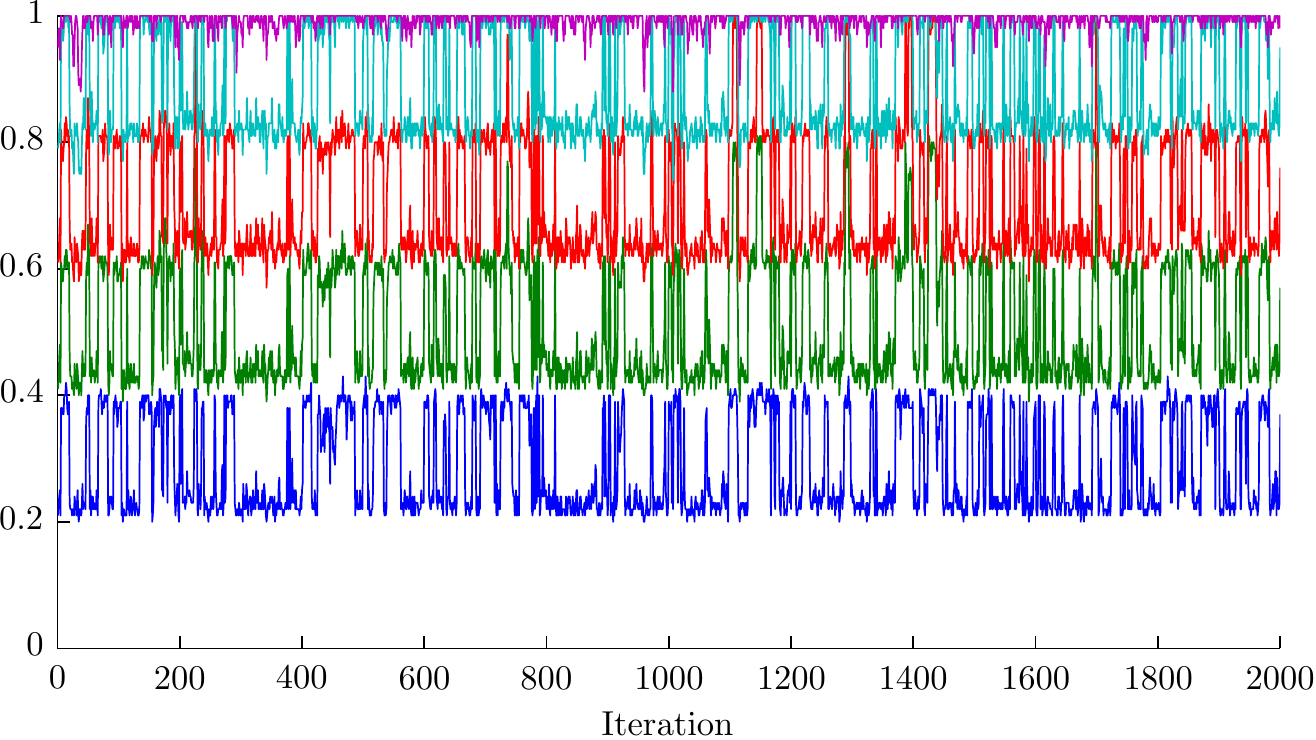}
                \caption{ESM}
                \label{fig:btrace3}
        \end{subfigure}
        \caption{Example trace plots of the fraction of observations in the largest, second largest, third largest etc. block as inferred using Gibbs, SM and ARM samplers. Data was generated from Example 1 of the Bernoulli Mixture Model. For the given simulation ARM obtained better mixing than the other two and this is consistent with the other simulations, however there were significant variability in the difficulty of the (stochastically generated) sampling problems owning to their small size.}\label{fig:btrace}
\end{figure}

As the number of features $d$ grow the observations assigned to the true blocks $k=1,2,3$ and $k=4,5$ become harder to distinguish from each other and we consider three experiments (Example 1, $d=6$, Example 2, $d=8$ and Example 3, $d=10$). Similar to \cite{jain2004split} we computed the trace plot by considering the fraction of vertices contained in the largest block, the fraction contained in the largest and second-largest blocks and so on. A typical result can be seen in figure \ref{fig:btrace} for the three methods evaluated on the same problem. It should be noted that while ARM typically produced more jagged trace plots than SM (which in terms produced more jagged trace plots than Gibbs), there was a significant variability for different randomly generated problems and for some problems the number of components remained fixed at for instance 4 or 5 for all methods.

To get quantitative results we computed correlation time both for the first element of the trace plot and for the indicator function $\delta_{z_i,z_j}$ for $5 \times 3$ observations randomly selected from the 5 planted blocks. In the later case we report the average of the \emph{maximum} of the autocorrelation time of the $\frac{1}{2}14\times 15$ pairs. Recall the autocorrelation time is defined as $1 + 2 \sum_{\tau=1}^m r(\tau)$ where $r(\tau)$ is the sample autocorrelation at the lag $\tau$ and the sum is terminated at values beyond which the autocorrelation is close to zero~\citep{neal1993probabilistic}.

To be consistent with \cite{jain2004split} we used iterations rather than wall-time to compute the autocorrelation. As can be seen the total computational effort of ARM and SM are roughly equally expensive on this problem while both are about 3 times more expensive than Gibbs sampling. All results are averaged across 20 different randomly generated data sets, each being sampled by $S=8$ different chains evaluated for $T=2000$ iterations using standard settings.

The results in table~\ref{table:jn} show quite large variations. This is mainly due to the stochastic nature of the generated data, however it makes comparison with \cite{jain2004split} difficult since those results are only based on a single simulation evaluated for half as many iterations and with autocorrelation computed between only a single pair of observations. Our results are however consistent with their conclusion in showing split-merge result in significantly lower autocorrelation times (though these need to be seen in the light of the higher computational effort) than Gibbs sampling while ARM seem to perform better than both methods on average.

To limit the effect of the variability in the data we plotted the estimated autocorrelation times found by ARM vs. those found using SM or Gibbs in a 2D scatter plot. To reduce the number of points we have shown the mean across the $S$ samples on the same data, see figure~\ref{fig:fig6jain_scatter}. This scatter plot illustrate the variability in the autocorrelation times and indicate significant improvement of ARM over the other methods.
\begin{table}
\centering
\begin{tabular}{l l  S  S  S  S }
\toprule
  { }   &  {Method}  &  \multicolumn{2}{c}{Autocorrelation}  &  {Normalized iterations}  \\ 
 \cmidrule(lr){3-4} 
 { } & { } & {Trace} & {Indicator} & { } \\ \midrule 
\multirow{3}{*}{Example 1} & Gibbs & 116.5 +- 93.9 & 91.6 +- 88.9 & 2000  \\  
  & SM & 27.8 +- 12.2 & 23.2 +- 10.3 & 6741 +- 468  \\  
  & ARM & 14.9 +- 7.7 & 13.6 +- 7.1 & 6712 +- 47  \\  \midrule  
 \multirow{3}{*}{Example 2} & Gibbs & 165.4 +- 165.6 & 131.3 +- 135.9 & 2000  \\  
  & SM & 27.9 +- 14.9 & 26.4 +- 14.9 & 6855 +- 482  \\  
  & ARM & 8.5 +- 5.0 & 9.2 +- 4.9 & 6731 +- 59  \\  \midrule  
 \multirow{3}{*}{Example 3} & Gibbs & 88.9 +- 137.4 & 99.3 +- 132.7 & 2000  \\  
  & SM & 19.3 +- 17.1 & 30.4 +- 25.4 & 7503 +- 185  \\  
  & ARM & 10.8 +- 12.5 & 14.7 +- 14.0 & 6705 +- 44  \\  
 \bottomrule
\end{tabular}

\caption{Artificial data simulation results for the Bernoulli Mixture Model. The methods was evaluated on the simulated data from Example 1-3. The ARM method find significantly lower autocorrelation time both for the trace plot of the fraction of observations in the largest block and as measured by co-occurrence of observations to the same blocks. The normalization implies the autocorrelation times can be compared directly. }\label{table:jn}
\end{table}


\subsection{Relational Modelling}\label{sec:IRMsimulations}
Our second example is the Infinite Relational Model (IRM) of \cite{Kemp06}, a non-parametric extension of a Potts-type spin model to the case of an unbounded number of partitions. For symmetric network data the generative process and log likelihood becomes
\begin{align*}
z & \sim \CRP(\alpha) &    \theta_{\ell m} & \sim \Beta(\beta_0^+, \beta_0^-), \ 1\leq m \leq \ell \\
A_{ij} & \sim \Bernoulli(\theta_{z_i z_j}) &
\mcal L(z) & = \sum_{1 \leq k<k' \leq K} \log\left( \frac{B(N^+_{kk'} + \beta_0^+, N^-_{kk'} + \beta_0^-)}{B(\beta_0^+,\beta_0^-)} \right)
\end{align*}
where $N^+_{kk'} = \sum_{i \in z(k), j \in z(k'), i \neq j}A_{ij}/2^{\delta_{kk'}}$, $N^-_{kk'} = \sum_{i \in z(k), j \in z(k'), i \neq j}(1-A_{ij})/2^{\delta_{kk'}}$.We again fixed all parameters to one. While this model is formally similar to the Bernoulli Mixture model the coupling of all components through $\theta$ make inference more challenging.

\subsection{Choice of data}

Consider a simple data set constructed by planting $K$ equally-sized communities in a network of $n=dK$ vertices, such that the edge-probability between edges inside a community is higher than the edge probability between edges in different communities. Assuming recovery is possible we expect the sampler to quickly find a number $K' < K$ communities and slowly split communities until the sampler converge at around $K$ communities. Since each of the $K'$ communities is (roughly) comprised of the union of one or more of the $K$ smaller communities we can expect split-merge moves to function well especially if the community structure is clearly defined. It was data with a well-defined partition structure which was considered in the previous section.

As another extreme, consider the case of a regular $D$-dimensional grid with a translation-invariant boundary. In this setting it is not unrealistic to assume the average size/number of communities will be roughly constant, and accordingly Split-Merge may have very low accept rate since it assumes both the number of components and their size change. Since the problem is translation invariant this will lead to poor mixing. While the grid provide a degenerate example, it has been shown under a wide range of conditions that Potts models on random graphs will be characterized by an exponential number of different, overlapping partitions of roughly the same likelihood and not the nested, well-defined partitions favorably to Split-Merge sampling \citep{borgs1999torpid}.

\newlength{\fx}
\setlength{\fx}{5cm}
\newlength{\fy}
\setlength{\fy}{2.5cm}
\begin{figure}
        \centering
        \begin{tikzpicture}[inner sep=0pt]
        \tikzstyle{every node}=[font=\small]
        \foreach \m [count = \xi] in {3,1,2}
            {
            \foreach \dd in {1,...,4}{
                 \node (n\dd\m) at (\fx * \xi ,-\fy * \dd) {\includegraphics[width=0.33\textwidth]{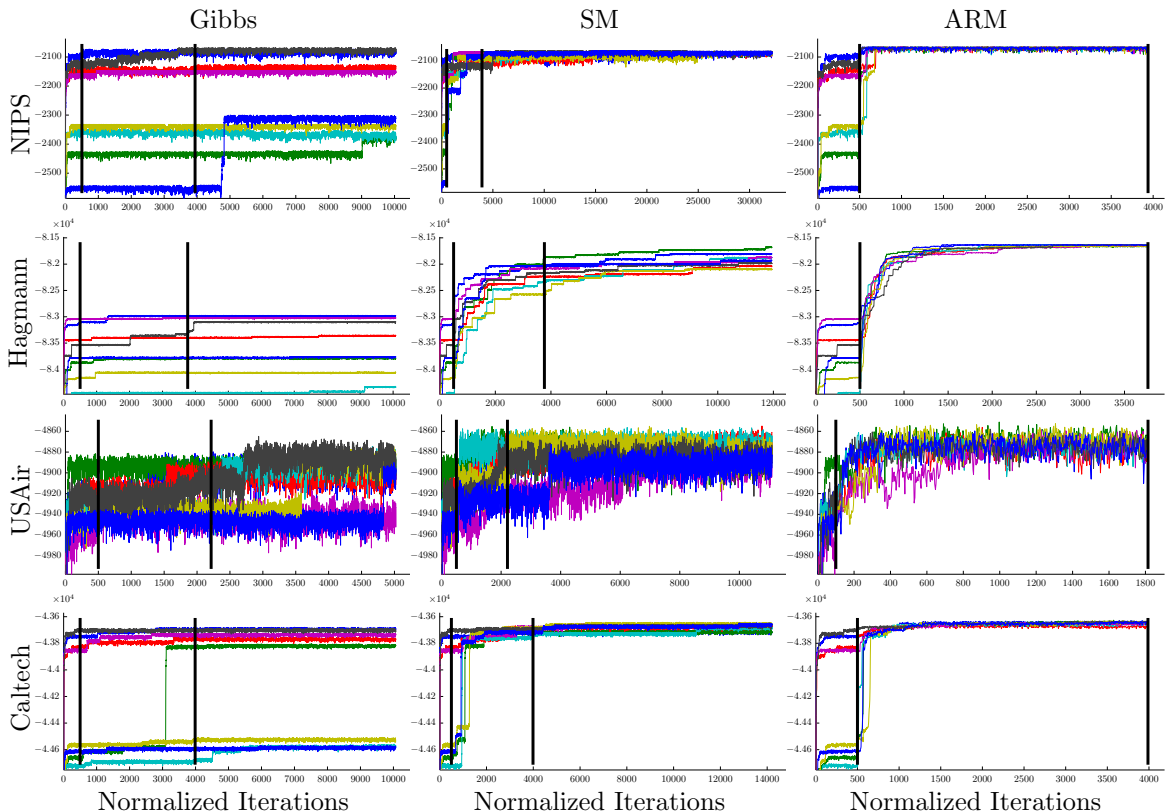}};
            }
        }
        \node[above of=n13,yshift=0.4cm] {Gibbs};
        \node[above of=n11,yshift=0.4cm] {SM};
        \node[above of=n12,yshift=0.4cm] {ARM};

        \node[left of=n13,rotate=90,yshift=1.7cm] {NIPS};
        \node[left of=n23,rotate=90,yshift=1.7cm] {Hagmann};
        \node[left of=n33,rotate=90,yshift=1.7cm] {USAir};
        \node[left of=n43,rotate=90,yshift=1.7cm] {Caltech};
        \node[below of=n41,yshift=-0.5cm] {Normalized Iterations};
        \node[below of=n42,yshift=-0.5cm] {Normalized Iterations};
        \node[below of=n43,yshift=-0.5cm] {Normalized Iterations};
        \end{tikzpicture}


        \caption{Trace plots of Log likelihood for Gibbs (left column), SM (middle column) and ARM sampling (right column) for all four network datasets.
         All simulations based on running $S=8$ chains using Gibbs sampling for 500 iterations, then continuing using Gibbs or SM for up to 10000 iterations or ARM for 1000 iterations. The $x$-scale is in normalized Gibbs iterations such that the space between the vertical black lines represent the same computational effort. The ARM method require significantly less effort to reach the same value of likelihood compared to the other methods.}\label{fig:fig1}
\end{figure}

In addition to these structures realistic network data may contain skewed degree distribution, core-periphery structure and small-world properties. It is difficult to assess the importance of these effects, and we therefore compare the sampling methods on real network data and use downsampling to adjust the difficulty of the sampling problem. The networks considered here is the \emph{NIPS} coauthor network ($n=234$) (available at \url{http://www.cs.nyu.edu/-roweis/data.html}), the USAir network ($n=332$) of US cities connected by flight routes in 1997 (available at \url{http://vlado.fmf.uni-lj.si/pub/networks/data}), \emph{Caltech} ($n=769$), the Caltech36 social network from the Facebook100 dataset (available at \url{http://datahub.io/dataset/facebook100}) and the \emph{Hagmann} structural brain network ($n=988$) of the number of fiber tracts between 998 brain regions as estimated by tractography from diffusion spectrum imaging across five subjects (see \cite{hagmann2008}).

All networks were prepared by symmetrizing adding the transpose, tresholding at $0$ and removing diagonal elements. To downsample a network from $n$ to $m$ vertices we sort the vertices according to the degree and label index (in descending order) and retain the $m$ first elements.


\subsection{Considerations for comparison}\label{sec:considerations}
Since ARM is about three times more expensive than a Gibbs sweep the raw number of iterations does not provide a fair point of comparison. A direct comparison of wall time showed high variance in the same problem due to differences in hardware and load, however, the cost of a single (full) Gibbs sweep can be considered equal for all samplers, and we therefore report the computational effort in units of Gibbs sweeps to obtain a standardized computational cost that allows for direct comparisons. For instance the number of standardized iterations for an ARM sampler evaluated for 1000 iterations was computed to be
$$
\text{Standardized Iterations} = 1000 \times \frac{\text{Total time spend}}{\text{Time spend on Gibbs iterations}}.
$$
As a further point the $S$ chains obtained by ARM are weakly dependent through the sharing of transition kernels, and comparison to a SM sampler which gives $S$ \emph{independent} estimates of the posterior at the same computational cost may be unfair. To account for this effect all comparisons between chains in the following sections is performed across different restarts and provide an unbiased estimate of the performance.



\subsection{Results - Full datasets}
To give a broad overview of the methods we evaluated Gibbs, SM and ARM on the full datasets and plotted the $log$ of the joint likelihood as a function of standardized iterations, see figure~ \ref{fig:fig1}. All simulations were obtained by running 8 separate chains with Gibbs sampling for 10000 iterations, and using the state of the chains after 500 iterations to initialize the SM chain (which was evaluated for an additional 9500 iterations) and the ARM sampler which was evaluated for an additional 1000 iterations. The vertical black lines are meant as a visual guide to indicate the same computational efforts.

The most striking feature is the consistent poor behavior of Gibbs sampling compared to the other sampling methods. When comparing ARM and SM sampling, notice even for the smallest network - the NIPS network of only 234 vertices, most SM chains only converge after many thousands of iterations whereas ARM converged after a few tens of iterations. This result is surprising since block-type models are routinely applied to problems much larger than the NIPS network. This behavior was robust across different restarts of both methods.

\begin{table}
\centering
\small
\sisetup{
table-figures-integer = 1,
table-figures-decimal = 2,
table-figures-uncertainty = 4,
table-number-alignment = center
}
\begin{tabular}{l c S S S S S S }
\toprule
& {Scale}  & \multicolumn{3}{c}{\mbox{GR-$\hat{R}$  ($T=2000$)} } & \multicolumn{3}{c}{\mbox{GR-$\hat{R}$  ($T=1000$)} }  \\ 
 \cmidrule(lr){3-5}\cmidrule(l){6-8} 
 &    & {Gibbs}  & {SM}  & {ARM}    & {Gibbs}  & {SM}  & {ARM}  
  \\ \midrule 
\multirow{3}{*}{NIPS}  & 1  & 20.46 +- 12.69   & 2.50 +- 1.53   & 1.04 +- 0.02   & 25.40 +- 10.90   & 3.24 +- 2.72   & 1.01 +- 0.00   \\
 & 0.9  & 23.28 +- 7.65   & 2.35 +- 0.74   & 1.00 +- 0.00   & 22.17 +- 7.64   & 2.74 +- 0.44   & 1.00 +- 0.00   \\
 & 0.8  & 11.22 +- 2.78   & 2.75 +- 1.11   & 1.01 +- 0.00   & 13.26 +- 4.52   & 2.89 +- 1.60   & 1.01 +- 0.01   \\
\midrule 
\multirow{3}{*}{Hagmann}  & 0.4  & 15.46 +- 13.63   & 4.24 +- 1.81   & 1.76 +- 0.05   & 15.96 +- 13.97   & 4.39 +- 2.37   & 1.77 +- 0.04   \\
 & 0.3  & 9.96 +- 6.78   & 1.69 +- 0.39   & 1.06 +- 0.07   & 10.62 +- 6.80   & 1.81 +- 0.46   & 1.13 +- 0.16   \\
 & 0.2  & 3.78 +- 2.39   & 1.02 +- 0.01   & 1.01 +- 0.00   & 4.54 +- 2.29   & 1.19 +- 0.23   & 1.02 +- 0.01   \\
\midrule 
\multirow{3}{*}{USAir}  & 0.9  & 2.49 +- 1.24   & 1.30 +- 0.29   & 1.02 +- 0.02   & 2.72 +- 1.00   & 1.82 +- 1.13   & 1.01 +- 0.00   \\
 & 0.8  & 2.41 +- 1.07   & 2.26 +- 0.99   & 1.00 +- 0.00   & 2.45 +- 1.13   & 2.19 +- 0.90   & 1.00 +- 0.00   \\
 & 0.7  & 1.04 +- 0.03   & 1.08 +- 0.04   & 1.02 +- 0.02   & 1.04 +- 0.03   & 1.12 +- 0.08   & 1.03 +- 0.02   \\
\midrule 
\multirow{3}{*}{Caltech}  & 0.8  & 32.13 +- 8.84   & 2.98 +- 1.01   & 1.18 +- 0.31   & 23.65 +- 16.66   & 2.37 +- 1.54   & 1.26 +- 0.32   \\
 & 0.7  & 4.32 +- 2.49   & 2.97 +- 1.44   & 1.01 +- 0.01   & 9.42 +- 9.40   & 2.71 +- 1.41   & 1.04 +- 0.03   \\
 & 0.6  & 8.49 +- 4.62   & 2.44 +- 0.76   & 1.01 +- 0.01   & 7.28 +- 3.35   & 2.56 +- 0.82   & 1.03 +- 0.02   \\
\bottomrule
\end{tabular}

\caption{Results for varying the downsampling (scale is the fraction of remaining observations) as well as the number of (normalized) iterations. Gelman-Rubin (GR) statistics are computed on the trace plot of the likelihood for $4$ different restarts and $S=8$. Half the samples are discarded as burning. Gelman-Rubin statistics less than 1.1-1.2 are normally considered compatible with mixing and ARM obtain lower GR values in fewer (normalized) iterations compared to the other methods. }\label{table:downsampling}
\end{table}

For the larger networks ARM obtain higher values of the likelihood more consistently than SM, sometimes with a significant margin as in the case of the Hagmann network, however, both methods did not converge except on the NIPS network. Notice, for Gibbs sampling each chain will behave in a stationary fashion for long stretches of time until it jump to a more favorable plateau. Since the chain appears to be converged during this time the most principal comparison seems to be between-chains statistics. We have focused on the Gelman-Rubin potential scale reduction factor $\hat{R}$ which attempt to quantify the likelihood samples from different chains came from the same distribution. Ideally $\hat{R}$ should be near 1, and values lover than 1.1 or 1.2 is generally considered consistent with convergence~\citep{brooks1998general}. We experimented on different quantities used to compute the GR statistics but settled on the joint log likelihood which seem to give a reasonable discriminative power. As described in section \ref{sec:considerations} we always compute GR statistics between different restarts creating an unbiased estimate and while we report simulation time in terms of number of ARM iterations we compute the statistics for the other method based on a similar number of normalized Gibbs iterations as described in section~\ref{sec:considerations}.

Simulation results on downsampled networks is available in table \ref{table:downsampling} for simulations where $S=8$ and the GR-statistics are computed based on $4$ restarts. Half the samples were discarded as burnin in the experiments.

Many of the networks required quite aggressive downsampling (the level was selected based on a coarse search as the highest value where at least one method have a favorable GR statistic) and the GR statistics show considerable variance. Somewhat counter-intuitively, a chain which is far from convergence will tend to have an increasing value of the likelihood during the sampling. Since this will inflate the within-chain variance, it will tend to lower the GR-statistics, and this effect seems partially responsible for the large variance. As can be seen none of the methods mix for the considered number of iterations except possibly the NIPS network. After downsampling Gibbs sampling continue to perform very poorly and ARM consistently perform better than SM sampling. Interestingly doubling the number of iterations from 1000 to 2000 does not seem to have a large effect.


\subsection{Effects of variation of the method}
The proposed method consist of several components such as the use of multiple parallel chains and the manner in which blocks of observations are reassigned to create bolder moves. To assess the impact of these ideas we will briefly discuss some variations of the method.

\subsubsection{Varying in the number of chains}
\begin{figure}
        \centering
        \begin{subfigure}[b]{0.5\textwidth}
                \includegraphics[width=\textwidth]{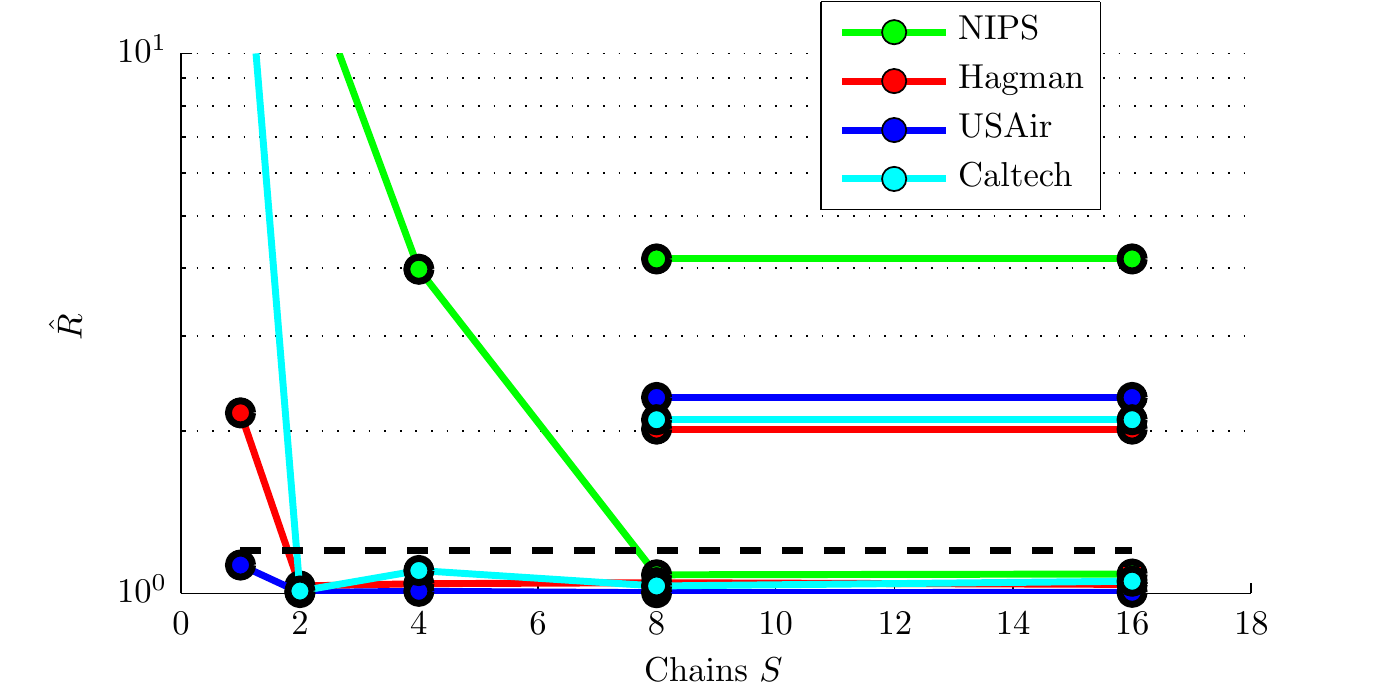}
                \caption{Varying number of chains}
                \label{fig:fig4chains}
        \end{subfigure}%
        ~ 
        \begin{subfigure}[b]{0.5\textwidth}
               \includegraphics[width=\textwidth]{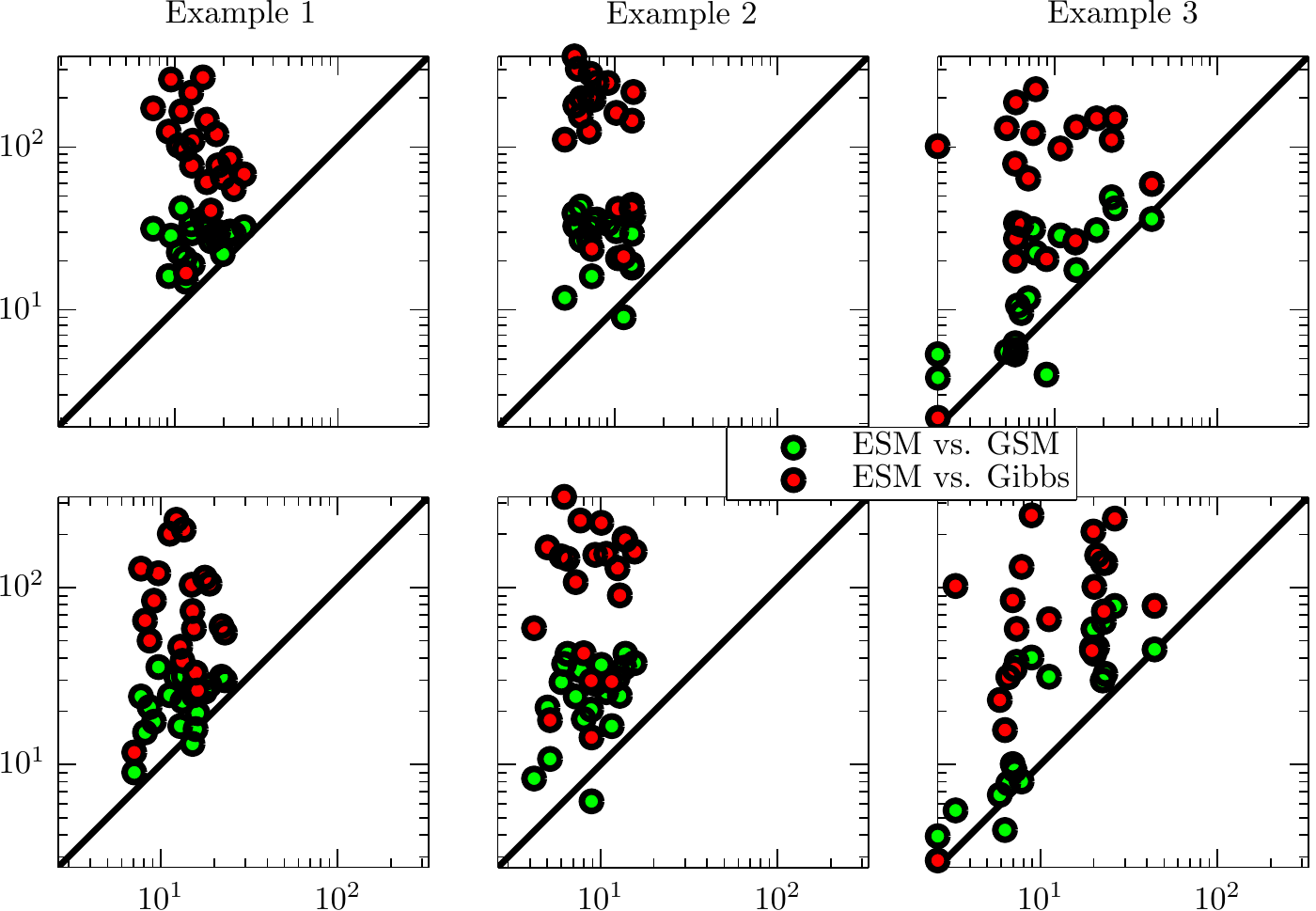}
                \caption{Autocorrelation plots}
                \label{fig:fig6jain_scatter}
        \end{subfigure}
        \caption{\emph{Left: } Effect of varying the number of chains on $\hat{R}$. For comparison we show results obtained using SM sampling as a vertical line. As can be seen, ARM benefit from the use of additional chains.
        \emph{ Right: } Scatter plots of autocorrelation time for the trace of the first block (top row) and for the indicators (bottom row) for the Bernoulli mixture model as described in the text. Points above the diagonal line indicate ARM perform better than the other method. As can be seen all methods have significant variances, however, there is a clear trend towards ARM performing better than SM for the considered problems.}
\end{figure}

An important question is how well the method benefit from improved parallelism. The simplest experiment is to vary the number of parallel chains $S$ between 2 and 16 for relevant levels of downsampling chosen from table~\ref{table:downsampling}. The result can be seen in figure~\ref{fig:fig4chains}.
As can be seen the method typically benefit from higher values of $S$ and for very small values of $S$ ARM sampling sometimes perform worse than Split-Merge sampling. This underperformance is not necessarily due to the method producing worse proposals since the additional computational cost of ARM means that it will consider significantly less proposal splits.

\subsubsection{Use of blocks and initialization}
To investigate the use of blocks when performing splits as well as how the method benefit from selecting the "correct" initial variables $i,j$ and initial splits, we proposed two variants of the SM and ARM methods. For the SM method, we choose the same initialization as in ARM, i.e., instead of randomly distributing the elements contained in $z_i \cup z_j \setminus \{i,j\}$ between the two new communities in the launch state we now select the same quintet $(i,j,z^a,z^b)$ as the ARM method and initialize the split configuration in the launch state in a similar way as ARM. Specifically, we set $z^{(l)}_i = c_i$ and $z^{(l)}_j = c_j$ where $c$ is the coarsest common refinement between $z,z^a$ and $z^b$. We dub this method bSM.

 \begin{table}
\centering
\sisetup{
table-figures-integer = 2,
table-figures-decimal = 2,
table-figures-uncertainty = 3,
table-number-alignment = center
}
\begin{tabular}{l l S S S S S}
\toprule
 & {Scale}   & {Gibbs}  & {SM}  & {bSM}  & {sARM}  & {ARM}  \\ 
    \midrule 
NIPS & 1  & 24.88 +- 8.23   & 2.43 +- 0.97   & 1.12 +- 0.15   & 2.41 +- 2.77   & 1.09 +- 0.03   \\
Hagmann & 0.25  & 2.86 +- 0.88   & 1.99 +- 0.84   & 1.65 +- 0.40   & 1.28 +- 0.33   & 1.08 +- 0.09   \\
USAir & 0.8  & 2.32 +- 1.44   & 1.94 +- 1.05   & 1.54 +- 0.74   & 1.60 +- 1.02   & 1.01 +- 0.01   \\
Caltech & 0.7  & 20.58 +- 13.96   & 3.41 +- 1.19   & 1.71 +- 1.19   & 1.61 +- 0.96   & 1.11 +- 0.09   \\
\bottomrule
\end{tabular}

\caption{Results for variations over SM and ARM samplers on downsampled network data. The numbers reported is the GR statistic of the log joint likelihood based on 1000 iterations (half discarded as burnin) for $S=8$ and 4 restarts. GR scores lower than 1.1-1.2 are usually considered consistent with mixing. The results indicate that while various features of the ARM sampler give some benefit individually the full method perform better than the variations considered individually.}\label{table:allmethods}
\end{table}

For the ARM sampler we considered a variant very similar to sRM described in algorithm~\ref{alg:ESMprime}. However, in order to create a fair comparison we still select $i,j$ and the initial partitions $z^a$,$z^b$ the same way as for regular ARM and again attempt the informed initialization where $z^*_i = c_i$ and $z^*_j = c_j$. This variant is dubbed sARM. Both new methods are interlaced with Gibbs sweeps and initialized similar to SM and ARM.

We compared these methods on downsampled networks for the equivalent of 1000 ARM iterations (half discarded as burnin) based on 4 restarts and $S=8$. The results can be seen in table~\ref{table:allmethods}.

Again ARM seem to perform either on par or better than SM, bSM and Gibbs while being slightly better than sARM, the magnitude seem quite dependent on the dataset. To better understand the behavior of the different methods we also computed the mean of the accept rate ($\times 100$) from eq.~\eqref{eqn:acceptrate} for all methods except Gibbs sampling. The clearest pattern is SM has accept rate less than half a percent for all datasets and the use of past states to guide the initialization only seem to double this number. While the accept rate of ARM is larger than for sARM on all datasets, it is much higher for the NIPS and Hagmann networks while being fairly low for USAir.

\section{Discussion}
Due to the explicit block-structure present in the Bernoulli mixture model examples we expected these to provide an ideal sampling problem for the split-merge method. While all samplers could easily solve the problem (computing GR statistics of the log joint likelihood gave results very near 1 in all instances), it seems that ARM and SM perform better than Gibbs sampling and ARM performs better still than SM.

These results are convolved by the high variability in autocorrelation times. Consider for instance the autocorrelation of the trace plot as seen in figure \ref{fig:btrace}. While Gibbs sampling no doubt has a high autocorrelation time, a problem where Gibbs sampling perform slightly worse and stayed in a single mode for the entire duration of the simulation would counterintuitive have a very low autocorrelation. This problem is aggravated by the stochastic nature of the test data and gave very high variance in the reported quantities. Still, since the autocorrelation is a bounded quantity, higher variance can be expected to be correlated with worse performance and there is a distinct trend towards ARM having lower mean autocorrelation and lower variance than SM. This conclusion is reinforced by figure~\ref{fig:fig6jain_scatter} which shows very high between-dataset variability but a clear trend towards most points being above the diagonal line indicating better performance by ARM.
\begin{table}
\centering
\begin{tabular}{l l S S S S}
\toprule
 & {Scale}   & {SM}  & {bSM}  & {sARM}  & {ARM}  \\ 
    \midrule 
NIPS & 1  & 0.48 +- 0.03   & 1.01 +- 0.08   & 1.37 +- 0.14   & 19.41 +- 1.28   \\
Hagmann & 0.25  & 0.14 +- 0.01   & 0.37 +- 0.06   & 1.08 +- 0.27   & 13.56 +- 2.42   \\
USAir & 0.8  & 0.04 +- 0.02   & 0.11 +- 0.04   & 0.35 +- 0.05   & 4.10 +- 0.19   \\
Caltech & 0.7  & 0.09 +- 0.04   & 0.32 +- 0.08   & 0.84 +- 0.09   & 6.82 +- 0.43   \\
\bottomrule
\end{tabular}

\caption{Acceptrate$\times 100$ for variations of SM and ARM samplers on downsampled network data. As can be seen, the variation in accept rate can be very dramatic when comparing SM to ARM. }\label{table:allmethods_accrate}
\end{table}

For the more realistic network-data examples table~\ref{table:allmethods_accrate} of averaged accept rates provide an interesting comparison between the methods. The most visible feature is the abysmal accept rate of split-merge sampling, typically a few tens of a percent to at most half a percent. It is interesting that despite such low accept rates split-merge sampling still provides improvements over Gibbs sampling for all datasets (see table~\ref{table:allmethods} and figure~\ref{fig:fig1}). The low accept rate is likely due to the nature of split-merge sampling requiring the number of components to grow or shrink by one, this view is supported in that the variant of SM sampling, bSM, which attempt to reuse past information of chains to provide the same initialization as ARM only has about twice the accept rate compared to the full ARM method which has accept rates up to 10-20\% for the NIPS and Hagmann network.

Focusing on the role of moving blocks, we see in particular for the networks with high accept rate (NIPS and Hagmann) movement of blocks cause the accept rate to increase by a factor 5-10 (compared to sARM). While the gains are more modest for the other problems they remain a consistent feature. As a corollary sARM outperform bSM in terms of accept rate uniformly. Taken together with the increased performance of ARM compared to SM, Gibbs, bSM and sARM in table~\ref{table:allmethods} this support the idea that moves that allow exchange of observations with blocks other than those presently being split or merged should be an important guiding feature in creating future samplers for partition-based problems, and movement of blocks boost the accept rate.

ARM seem to perform better than SM in nearly all settings, see table~\ref{table:downsampling}, and for most datasets the increase in performance does not seem to be easily matched by simply increasing the simulation time of SM. The difference in behavior is perhaps most striking when simply visualizing the trace plots (see figure~ \ref{fig:fig1}) since when translating the performance into numerical quantities the methods which has not converged show high variance. However, the trace plots indicate that SM can be considered tens of times slower than ARM on realistic datasets.

The use of past states not only has a positive effect on the accept rate but also help exploration. This is indicated in figure~\ref{fig:fig4chains} which suggests the minimum number of parallel chains to be used should be about 4-8, but that the method seem to scale favorably with increased parallelism. This is a significant result in light of the current trend where computers scale in parallelism rather than speed.

\section{Conclusion and Further Research}
We have presented a novel method for sampling partition-based models. The proposed method evaluates an ensemble of chains in parallel and use their current and past states to construct an adaptive transition kernel. While the transitions will either split or merge two selected observations, in contrast to standard Split-Merge operations the number of blocks may increase, decrease or stay the same regardless of the type of operations.

Our simulations indicate the method has superior performance over Split-Merge or Gibbs sampling for the Bernoulli Mixture model and the Infinite Relational Model on both artificial and real data, and by considering variants of the method we have shown the particular use of past states as well as the extended space of available transitions both contribute towards the methods performance, and the use of multiple chains play a crucial role in providing exploration and increased accept rate. The increased parallelism in modern computing methods can take advantage of this capacity.


It is worth emphasizing the two key components in the present method, the use of past states to inform the current moves and the use of bolder proposals can likely be furthered with some modifications to the above framework. Some proposals for future research could be softening the requirement that only two observations, $i,j$, are forcibly split or merged, examining the role of using more than two past states in constructing the transition kernel or allowing movement of blocks of variables from outside the initial states $i,j$.

The extension to non-conjugate models is fairly straight forward using techniques such as Algorithm 8 of \cite{neal2000markov}. However, a procedures more in line with the current work could be to use the value of the (non-conjugate) parameters obtained from the initial states $z^a$ and $z^b$ rather than draws from the prior to inform the proposal distributions.

Despite the wide application of partition-based Bayesian models the construction of good transition kernels remain a largely under-explored area. Our results suggest transitions involving more than two blocks are of vital importance to obtain high accept rates, adding to the problem of how to propose near-equilibrium states. Our current proposal attempts to alleviate the later issue using information from past states rather than restricted Split-Merge operations. While there is no doubt many variations of how this information could be gathered or put the \emph{need} to include such information in the proposal distribution seem to be a robust feature. 

\acks{This project was funded in part by the Lundbeck Foundation.}

\vskip 0.2in
\bibliographystyle{plainnat}
\bibliography{library}
\end{document}